\def\year{2018}\relax
\documentclass[letterpaper]{article} 
\usepackage{aaai18}  
\usepackage{times}  
\usepackage{helvet}  
\usepackage{courier}  
\usepackage{url}  
\usepackage{graphicx}  
\usepackage{color}
\usepackage[utf8]{inputenc} 
\usepackage[T1]{fontenc}    
\usepackage{hyperref}       
\usepackage{booktabs}       
\usepackage{amsfonts}       
\usepackage{nicefrac}       
\usepackage{tabularx}
\usepackage{tabu}
\usepackage{booktabs}
\usepackage{epsfig}
\usepackage{amsmath}
\usepackage[square,numbers]{natbib}

\begin{document}
\title{Deep Multiple Instance Feature Learning \\ via Variational Autoencoder}
\author{
Shabnam Ghaffarzadegan \\
Bosch Research and Technology Center, Palo Alto, CA 94304, USA \\
\texttt{shabnam.ghaffarzadegan@us.bosch.com}}
  
\maketitle
\begin{abstract}
We describe a novel weakly supervised deep learning framework that combines both the discriminative and generative models to learn meaningful representation in the multiple instance learning (MIL) setting. MIL is a weakly supervised learning problem where labels are associated with groups of instances (referred as bags) instead of individual instances. To address the essential challenge in MIL problems raised from the uncertainty of positive instances label, we use a discriminative model regularized by variational autoencoders (VAEs) to maximize the differences between latent representations of all instances and negative instances. As a result, the hidden layer of the variational autoencoder learns meaningful representation. This representation can effectively be used for MIL problems as illustrated by better performance on the standard benchmark datasets comparing to the state-of-the-art approaches. More importantly, unlike most related studies, the proposed framework can be easily scaled to large dataset problems, as illustrated by the audio event detection and segmentation task. Visualization also confirms the effectiveness of the latent representation in discriminating positive and negative classes.

\end{abstract}

\section{Introduction}
\noindent Applications of machine learning usually require accurately labeled training data. Recent remarkable breakthroughs in deep learning made this requirement even more crucial, where large amount of carefully annotated data is required to train complicated networks \cite{Hinton:2006:FLA:1161603.1161605, lecun2015deep}. However, creating the labeled data usually involves human annotation that are associated with high cost and potential human errors\cite{krizhevsky2012imagenet}. One way to relax this constraint is using the multiple-instance learning (MIL) framework\cite{DIETTERICH199731}. Unlike traditional supervised learning where training data includes instance (feature) and label pair; for MIL, each training example consists of a group of instances (referred as bag) and the associated label. In the binary classification setting, a bag is labeled as negative when all the instances in the bag are negative instances. On the contrary, a bag is labeled as positive when at least one of the instances in the bag is a positive instance. The MIL setup relaxes the data annotation requirement by allowing ambiguity in the labels of the positive bag instances. Take image object recognition for example, instead of relying on accurate boundary of the object-of-interest, MIL directly uses the label of the image by considering multiple patches in the image as a bag where each patch may or may not include the object of interest. Another example is audio event detection, instead of having precise boundaries of an audio event of interest which is time consuming and expensive, MIL can rely on a coarse label of a windowed audio signal segment that contains the audio event. 

The flexibility on the requirement of data annotation comes with the cost of increased difficulty in the learning tasks. This is mainly due to the ambiguity in the positive instance labels where the positive bag could contain both negative and positive instances. Directly using traditional machine learning approaches with the bag level label fails to consider the incorrect label of the negative instances in the positive bag. As a result, previous studies have shown that MIL specific algorithms performs better in this setting  \cite{babenkomultiple}. While previous studies mostly focus on learning decision boundaries using the bag level representations, such as Diverse Density\cite{Maron98}, Citation-KNN\cite{Wang:2000:SMP:645529.757771}, Mi-Graph\cite{Zhou:2009:MLT:1553374.1553534} and SVM based approaches\cite{NIPS2002_2232,DBLP:journals/corr/WangZYB15}. There is relatively little work on characterizing the optimal representation of the data in this setting.

In this paper, we present an approach that combines the discriminative and generative models to learn meaningful low-dimensional feature representation in the MIL setting. Given a set of training bag data, we train two generative models to learn a latent representation of the negative instances and all instances using variational autoencoder \cite{kingma2013auto}, noted as $VAE_{-}$ and $VAE_{\pm}$, respectively. We use the latent representations of the $VAE_{\pm}$ to distinguish the positive bag instances from the negative bag instances using a discriminative classifier. To take into account of the uncertainty in the positive bag instances, we apply weighted loss to the positive bag instances based on their reconstruction error from $VAE_{-}$. Intuitively, if an instance in the positive bag is very similar to the negative bag instances, it is likely to be the negative instance and should not be penalized much for being classified as positive instance. By simultaneously training these three components, a low-dimensional latent representation of $VAE_{\pm}$ are optimized to capture the difference between the positive and negative instances while preserving the characteristics of the individual class. Using our proposed approach, we observe superior performance across different MIL benchmark datasets and the audio event detection tasks.

The main contribution of this work is to demonstrate how to incorporate generative model, VAE in this case, into MIL problem in a principled manners. The proposed framework can be regarded as a discriminative model regularized with VAEs. It explores the unique challenge in MIL setting where positive instance labels are ambiguous. More importantly, the fixed-size low dimension latent representation enables the proposed framework to be applied to large dataset with high dimensional features, where most other related studies fail to apply due to high computational complexity. The remainder of this paper introduces our method and experiments in detail.

\subsection{Related Work}
\textit{Learning Axis-Parallel Concepts}: Learning Axis-Parallel Concepts, are the  first group of methods used by \cite{DIETTERICH199731, Auer:1997:AHL:258533.258611, Long1998} to solve the MIL problem. In these methods, the goal is to find an axis-parallel hyper-rectangle (APR) in the feature space to represent the target concept. However, this methods lack of practical application since they neglect majority of data in large bag sample\cite{babenkomultiple}.

\textit{Maximum Likelihood}: Similar to their counterparts in traditional machine learning setting, the maximum likelihood in MIL aims to train a classifier that maximize the likelihood of the data. Many methods approximate a differentiable loss function to perform gradient descent \cite{Ramon00multiinstance, babenko:inria-00326736, NIPS2005_2926}. The most well-known method of this category is Diversity Density (DD) proposed by \cite{Maron98} as a general multi-instance learning framework where each bag is regarded as a manifold composed from many instances. In this method the classifier consists of only one vector from the input space called the target point, which is a data point that is close to at least one instance from positive bags while being far from instances in negative bags, known as diversity density measure. Zhang \textit{et al.} \cite{Zhang01em-dd:an} proposed an Expectation-Maximization based DD method (EM-DD) that iterates over two steps: in \textit{E step} the current classifier is used to choose the most probable point from each positive bag, and in \textit{M step} standard supervised learning method is used to find a new concept point by maximizing likelihood over all the negative and positive instances. This concept was further developed by many other related MIL methods \cite{NIPS2002_2232}.
 
\textit{Maximum Margin}: Maximum margin method used in support vector machine (SVM) can be adapted to MIL framework. Andrews \textit{et al.} \cite{NIPS2002_2232} proposed two different methods namely mi-SVM and MI-SVM, for instance-level and bag-level classification respectively, to define the margin for positive bags. Both algorithms follow similar iteration process as in EM-DD method. In MI-SVM only one positive instance in each positive bag contributes to the optimization and the other instances in positive bags are ignored. By contrast, mi-SVM method considers both positive and negative instances in the positive bags while optimizing the support vectors. Following these two ideas, many different variations of SVM based methods for MIL task are developed\cite{DBLP:journals/corr/WangZYB15, Chen:2004:ICL:1005332.1016789}.

\textit{Deep learning based}: Recent developments in deep neural networks are also applied to the MIL problems with the assumption that meaningful features can be learned directly from the bag level labels by the network. \cite{DBLP:journals/corr/SongLJD14} used convolutional neural network (CNN) for feature learning on a weakly supervised object localization task. In \cite{DBLP:conf/icassp/XuMFZLC14}, deep neural network is used to learn features for weakly supervised learning in medical imaging. \cite{conf/cvpr/WuYHY15} reformulated the CNN loss function to train an end-to-end solution for image classification and annotation problems. In \cite{feng2017deep}, Feng \textit{ et al.} took the challenge to solve multiple instance multiple label (MIML) problem by proposing the matching score between the instance and sub labels. In this setting, the additional information among multiple labels can be used to facilitate bettering learning. In \cite{wei2017scalable}, the authors solve MIL problem by applying two different hand-engineered feature representations, including locally aggregated descriptors and Fisher vector, to convert the bag into a fix size vector representation.

Our approach combines the ideas behind the Maximum Likelihood, Maximum Margin and deep learning based approaches. It aims to minimize the distance between negative instances, while separating the positive instances as far as possible through learning a new representation of the data. Instead of relying on single instance distance based measurement, such as in DD and MI-SVM, the proposed framework is built on instance distribution learned from the variational autoencoder. This framework allows us to address the uncertainty in the positive label by measuring the similarity between the instance in the positive bags and the negative instances in terms of reconstruction error in the VAE trained with negative instance only. The effectiveness of the proposed approach is illustrated by comparing the performance on different MIL benchmark dataset with previous state of art approaches. More importantly, by taking the advantage of deep learning training mechanism, the proposed framework provide a scalable solution in contrast to most of the previous MIL research which are not capable of handling large scale datasets \cite{DBLP:conf/icdm/WeiWZ14}. In fact, this issue was previously addressed by Wei \textit{et al.} in \cite{wei2017scalable} where they applied hand-engineered feature representations, including locally aggregated descriptors and Fisher vector, to convert the bags into a low-dimension fix size vector representation allowing fast learning. However, these hand-engineered feature may not achieve the best possible representation of the bags thus lead to sub-optimal performance. In this work, we train the network to automatically explore a meaningful representation and achieve better performance in comparison. These advantages are discussed in detail in the experiment sections.

\begin{figure*}[t] 
 \centering
 \epsfxsize=140mm \epsffile{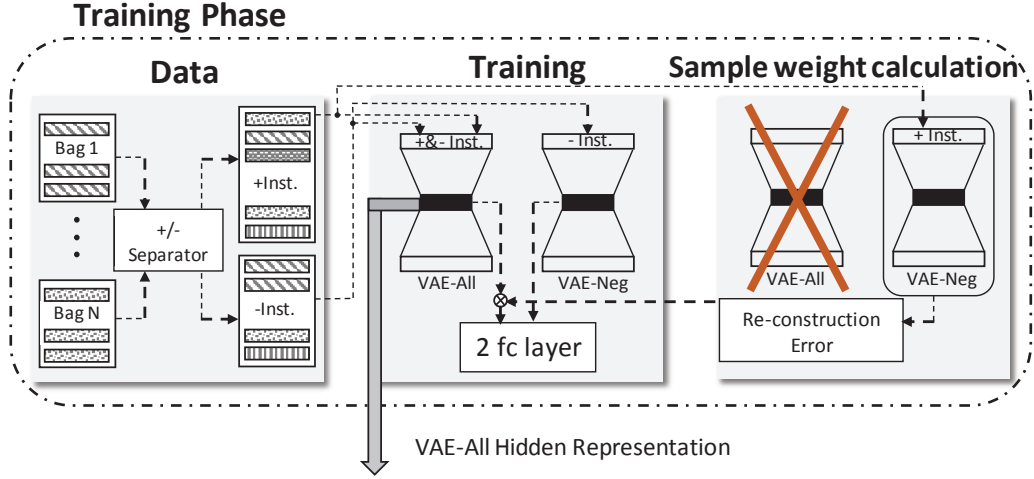}
 \caption{\footnotesize We propose a Variational Autoencoder network for learning feature representation in the MIL setting. In the training phase, pair-wise samples are fed to two VAEs ($VAE_{\pm}$ and $VAE_{-}$) to learn two posteriors $p(z|X)$ and $p(z|X,Y=-1)$, respectively. The latent layers of both VAEs are concatenated to a classifier to determine whether both input samples are negative instances. Separator: to separate positive and negative bags based on training labels. \textit{fc}: fully connected. \label{fig:sys}}
\end{figure*} 

\section{Multiple Instance Feature Learning}

\subsection{Problem Setup}
Let $\mathbb{X}$ denote the feature space and $\mathbb{Y}$ denote the set of class labels. In MIL, the training set of $n$ examples is noted as $(\mathrm{X},\mathrm{Y}) = \left \{ (X_1,Y_1),\dots,(X_n,Y_n) \right \}$. Each example consists of a bag of $m_i$ instances $X_i=\left \{ x_{i1}, \dots, x_{im_i}  \right \}$ and the bag label $Y_i$ where $x_{ij} \in \mathbb{X}$ and $Y_i \in \mathbb{Y} = \left \{-1,1 \right \}$. A bag label $Y_i = -1$ if instance label $y_{ij}=-1$ for all $x_{ij} \in X_i$; Otherwise, $Y_i=1$. In other words, a bag is labeled as negative bag if all instances belong to the negative class; a bag is labeled as positive bag if at least one instance in the bag belongs to the positive class. Similar to the traditional machine learning, the goal of MIL is to learn a mapping function $\mathit{f} : X_i \rightarrow Y$ that minimizes a loss function $\mathit{l}$: $Y \times Y \rightarrow \mathbb{R}$. In this setting, a positive bag may contain negative instances thus learning directly from the bag level label carries the intrinsic error from the label. 

We are interested in learning a better feature representation that explores the difference between the positive and negative bags at the instance level. Given that the negative instances have correct labels and the positive instances have uncertain labels, a meaningful representation should consist of similar encodings among the negative instances, while encouraging different yet possible similar encodings between the positive and negative instances. Following this direction, we use two separate variational encoders to model two conditional distributions, namely all instances $p(z|X)$ and negative instances $p(z|X,Y=-1)$. Here $z\in\mathbb{R}^{n_{z}}$ represents the latent variables with a prior of $p(z)$. These latent representations are also used to explore the difference between the positive and negative bags through a discriminator (see Figure \ref{fig:sys}). These two training objectives encourages learning meaningful representations that not only encode the corresponding input data, but also implicitly capture the differences between the positive and negative instances. 

It is worth noticing that maximizing the difference between $p(z|X)$ and $p(z|X,Y=-1)$ is equivalent to maximizing the difference between $p(z|X,Y=1)$ and $p(z|X,Y=-1)$, which is the difference between the representation between positive and negative bags. Let $p_Y = p(Y=1)$ and $p(Y=-1)=1-p_Y$ being the prior of $p(Y)$, we have 

\begin{equation}
\begin{aligned}
p(z|X) = & p(z|X,Y=1)*p_Y+p(z|X,Y=-1)*(1-p_Y) \\ 
 = & p(z|X,Y=-1) + [p(z|X,Y=1)\\
 & -p(z|X,Y=-1)]*p_Y
\end{aligned}
\label{eq:1}
\end{equation}

It is easy to see the following equation holds.

\begin{equation}
\begin{aligned}
p(z|X) - &p(z|X,Y=-1)=\\
&[p(z|X,Y=1)-p(z|X,Y=-1)]*p_Y
\end{aligned}
\label{eq:2}
\end{equation}

\subsection{Variational Autoencoder}

There are many different approaches for learning latent representation of $p(z|X)$ and $p(z|X,Y=-1)$, we use the Variational Autoencoder (VAE) \cite{kingma2013auto} since it can be combined with the discriminator model for the MIL problem in a principled manner, as discussed in the next section. VAE is a deep directed graphical model consisting of an encoder and decoder. The encoder maps the data sample to a latent representation $p(z|X)$ and the decoder maps the latent representation back to the data space $p(X|z)$. The loss function of the VAE is defined as following:
\begin{equation}
	\mathcal{L}_{VAE} = \mathrm{KL}(q(z|X)\parallel p(z))-\mathbb{E}_{q(z|X)}[\log p(X|z)]
\label{eq:vae_loss}
\end{equation}
By regularizing the encoder with a prior over the latent representation $p(z)$, $z\sim \mathcal{N}(0,\mathbf{I})$ where $\mathbf{I}$ is identity matrix, the VAE learns a latent distribution $q(z|X)$ that contains sufficiently diverse representation of the data. 

\subsection{MIL Feature Learning Network}

Figure \ref{fig:sys} presents the proposed MIL feature learning network, which consists of two VAEs sharing the same configurations, and a classifier network that take the latent layer in VAEs as inputs. The two VAE networks approximate the posterior of $p(z|X)$ and $p(z|X,Y=-1)$, noted as $VAE_{\pm}$ and $VAE_{-}$ respectively. This is achieved by training the $VAE_{\pm}$ with all instances from both positive and negative bag examples, while training the $VAE_{-}$ with instances from only negative bag samples. By concatenating the latent representations to a discriminator to differentiate the positive instances from the negative instances, the overall network simultaneously optimize the latent representation and classification learning. The loss of the proposed network consists of $\mathcal{L}_{VAE_{\pm}}$, $\mathcal{L}_{VAE_{-}}$ and the binary cross-entropy loss for classifier $\mathcal{L}_{clf}$. To address the uncertainty of the positive instance label, we use the reconstruction error from the $VAE_{-}$ as sample weight for the  classifier loss $\mathcal{L}_{clf}$. The idea is that if an input instance in the positive bag can be well reconstructed by the $VAE_{-}$, it is likely to be a negative instance mislabeled by the positive bag label. Table \ref{tab:layers} shows the network configuration in Figure \ref{fig:sys}.

\begin{table}
\centering
\caption{Architectures for the VAE - number of nodes/layer structure/activation function. fc stands for fully-connected layer; $n_z$ represents VAE hidden layer size.}
\resizebox{\columnwidth}{!}{%
\begin{tabular}{c|c|c} \hline
\textbf{VAE Encoder}&\textbf{VAE Decoder}&\textbf{Classifier}\\ \hline
512 fc, ReLU& 256 fc, ReLU& 64 fc, ReLU\\
256 fc, ReLU& 512 fc, ReLU& 64 fc, ReLU\\
$n_z$ fc, ReLU & fc, sigmoid&2 fc, Softmax\\
\hline
\end{tabular}
}
\label{tab:layers}
\end{table}

The proposed framework applies the VAEs to the MIL problem in a principled manners. Let $\lambda_{\pm}$ and $\lambda_{-}$ be the parameters of $VAE_{\pm}$ and $VAE_{-}$ respectively. Given the prior of $z\sim \mathcal{N}(0,\mathbf{I})$, the $\mathrm{KL}(q(z|X)\parallel p(z))$ in the $\mathcal{L}_{VAE}$ encourages both $q_{\lambda_{-}}$and $q_{\lambda_{\pm}}$ to follow Gaussian distributions. We note them as $\mathcal{N}(\mu_{-},\Sigma_{-})$ and $\mathcal{N}(\mu_{\pm},\Sigma_{\pm})$ respectively. When we use the latent representation from $VAE_{\pm}$ and $VAE_{-}$ to distinguish the positive bag instances from the negative bag instances, we are indeed solving an optimization problem of $\lambda_{-}$ and $\lambda_{\pm}$ such that

\begin{itemize}
\item $q_{\lambda_{-}}$ and $q_{\lambda_{\pm}}$ estimates the posterior well.
\item The difference between $q_{\lambda_{-}}$ and $q_{\lambda_{\pm}}$
is maximized.
\end{itemize}

While the objective of VAE network is aligned with the first goal of this optimization problem, we notice that the second part of this problem can indeed be achieved by maximizing the difference between $\mu_{-}$ and $\mu_{\pm}$. Notice that the Kullback-Leibler (KL) divergence from $q_{\lambda_{-}}$ to $q_{\lambda_{\pm}}$ is 

\begin{equation}
\begin{aligned}
KL(q_{\lambda_{\pm}}\|q_{\lambda_{-}}) = & \\ \frac{1}{2}
  & \{\log\frac{|\Sigma_{-}|}{|\Sigma_{\pm}|}-n_{z}+
 {\rm tr}\left(\Sigma_{-}^{-1}\Sigma_{\pm}\right)+\\
 & (\mu_{-}-\mu_{\pm})^{\top}\Sigma_{-}^{-1}(\mu_{-}-\mu_{\pm})\}.
\end{aligned}
\end{equation}

\noindent Since $\Sigma_{-} \approx \mathbf{I}$ and $\Sigma_{\pm} \approx \mathbf{I}$,

\begin{equation}
\begin{aligned}
KL(q_{\lambda_{\pm}}\|q_{\lambda_{-}}) & =\frac{1}{2}\left[0-n_{z}+n_{z}+(\mu_{-}-\mu_{\pm})^{\top}(\mu_{-}-\mu_{\pm})\right]\\
 & =\frac{1}{2}\|\mu_{-}-\mu_{\pm}\|_{2}^{2}.
\end{aligned}
\end{equation}

While we could use a variety of distance metrics between the latent layers of $\mu_{-}$ and $\mu_{\pm}$ to maximize the difference between $q_{\lambda_{-}}$ and $q_{\lambda_{\pm}}$, such strategies may unnecessarily constrains the representation of the latent variables. Following the strategy similar to the idea of the Generative adversarial networks (GAN) \cite{NIPS2014_5423}, we instead use a classifier network that takes the latent layers of $\mu_{-}$ and $\mu_{\pm}$ as input to distinguish the positive bag instances from negative bag instances. In this way, the optimal distance metric is learned from the data.

\subsection{Training Detail}

During learning, we aim to use  $\mathcal{L}_{clf}$ to maximize the difference between the two posterior estimates $q_{\lambda_{\pm}}\|q_{\lambda_{-}}$ and $\mathcal{L}_{VAE_{\pm}}$, $\mathcal{L}_{VAE_{-}}$ to train a optimal posteriors of $p(z|X)$ and $p(z|X,Y=-1)$. Training data is prepared in pairs with the input to $VAE_{\pm}$ being randomly chosen from all instance, and input to $VAE_{-}$ being randomly chosen from the negative bag instances only. For robustness, we follow the data augmentation procedure similar to the concept introduced in \cite{doran2013smile} to repeat the aforementioned procedure multiple times. As a result, different positive-negative and negative-negative instance pairs will be included during the training. 

Our approach is implemented in Keras. We use the RMSprop optimizer and a initial learning rate of 0.001 and momentum of 0.9 throughout our experiments. We initialized all the weights to zero mean Gaussian noise with a standard deviation of 0.01. The code will be released at time of publication.

\subsection{Bag Level Classification}

\begin{figure}[t] 
 \centering
 \epsfxsize=80mm \epsffile{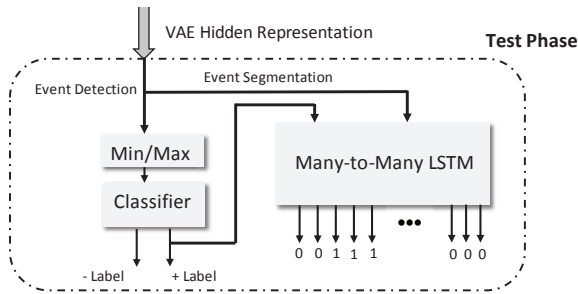}
\caption{\footnotesize Bag level classification and audio segmentation. Simple statistics are extracted from the latent representation learned from from $VAE_{\pm}$ and used for classification. For audio segmentation task, only the positive bags from the classification result with their VAE latent representation are fed to the LSTM to detect the start and end time of the event. \label{fig:sys_test}}
\end{figure}

To leverage the learned representation of the MIL data for binary  event detection task, we extract simple features from the encoded space including maximum, minimum, mean and standard deviation of the encoding value along each latent dimension as bag level feature $\in \mathbb{R}^{n_z \times 4}$. This feature representation is evaluated using 3 simple classifiers including k-nearest-neighbor classifier (KNN), neural network (NN) and Adaboost to show the effectiveness of the proposed framework. To detect the event boundaries, we use the encoded features as an input to a many-to-many long short term memory (LSTM) network. Figure \ref{fig:sys_test} illustrates the process.

\section{Experiment}
\subsection{Datasets} \label{data}
\textit{MUSK dataset}: The MUSK datasets introduced in \cite{DIETTERICH199731} have been used in all the previous MIL research as the benchmark sets. This data contain two sets of MUSK1 and MUSK2 with 166 feature vectors describing molecules using multiple low-energy conformations. MUSK1 consists of 92 molecules (bags) and average of 6 conformation per molecules (instances), and MUSK2 is composed of 102 molecules with average of 64.7 conformations per molecule. Moreover, MUSK1 dataset includes 479 instances divided into 47 positive and 45 negative bags; and MUSK2 contains 6600 instances partitioned into 39 positive and 63 negative bags.

\begin{table} [t] 
  \caption{Low level descriptive features and high level features (functionals) computed on audio data; \textit{min}: minimum; \textit{max}: maximum; \textit{std}: standard deviation; \textit{var}: variance; \textit{dim}: dimension}
  \label{audio_table}
\centering
  \begin{tabular}{c|c}
    \hline
     Features & Functionals  \\
    \hline
     {}& Min  \\
     & Max  \\
    Zero crossing rate \& $\Delta$ (2-dim) & std  \\
    Energy \& $\Delta$ (2-dim) & var  \\
    Spectral centroid \& $\Delta$ (2-dim) & skew  \\ 
    Pitch \& $\Delta$ (2-dim) & kurtosis \\
    MFCC \& $\Delta$ (26-dim) & mean \\
    {} & median \\
    \hline
  \end{tabular}
\end{table}

\textit{Automatic Image Annotation dataset}: Automatic image annotation assigns keywords to an image based on the context information. It can be formulated as MIL problem where each image is regarded as a bag where features of image patches are the instances. The benchmark datasets for image annotation include Tiger, Fox and Elephant datasets introduced in \cite{NIPS2002_2232}. They are extracted from Corel dataset \cite{MILES}. Each of these sets have 100 positive and 100 negative bags, in which positive bags correspond to images of the target animal and negative bags include images drawn randomly from the pool of other animals. In each bag, instances are created by segmenting the image and using color, texture and shape features as segment descriptors.

\begin{table*}[t]
  \caption{Average prediction accuracy (\%) using 10-
  cross validation on benchmarks. Some standard deviations are not provided by former studies.}
  \label{tab:benchmark_results}
  \centering
  \begin{tabular} {c|c|c|c|c|c}
    \toprule
    Method & MUSK1 & MUSK2 & Fox & Tiger & Elephant\\
    \hline
    \hline
     mi-SVM \cite{NIPS2002_2232} & 87.4 & 83.6 & 58.2 & 78.4 & 82.2 \\
     MI-SVM \cite{NIPS2002_2232} & 77.9 & 84.3 & 57.8 & 84.0 & 81.4 \\
     RMI-SVM\cite{DBLP:journals/corr/WangZYB15}& 80.8 & 82.4 & 63.6$\pm$2.8 & \textbf{87.9$\pm$0.9} & \textbf{87.8$\pm$0.7} \\
     EM-DD \cite{Zhang01em-dd:an} & 84.8 & 84.9 & 56.1 & 72.1 & 78.3 \\ 
     mi-Graph\cite{Zhou:2009:MLT:1553374.1553534}&88.9$\pm$3.3&90.3$\pm$2.6 &61.6$\pm$2.8&86.0$\pm$1.6&86.8$\pm$0.7 \\
     MI-Graph \cite{Zhou:2009:MLT:1553374.1553534} & 90.0$\pm$3.8 &\textbf{ 90.9$\pm$2.7} & 61.2$\pm$1.7 & 81.9$\pm$1.5 & 85.1$\pm$2.8 \\ 
     MI-Forests \cite{Leistner:2010:MML:1888212.1888216} & 85 & 82 & \textbf{64} & 82 & 84 \\ 
     DMIL\cite{conf/cvpr/WuYHY15} & 87.5 & 72.5 & 62.5 & 79.4 & 82.5 \\  
     MiFV\cite{conf/cvpr/WuYHY15} & 90.9$\pm$8.9 & 88.4$\pm$9.4 & 62.1$\pm$10.2 & 81.3$\pm$8.3 & 87.1$\pm$7.3 \\
     MiV\&F\cite{conf/cvpr/WuYHY15} & \textbf{91.5$\pm$8.3} & 88.1$\pm$8.7 & 62.0$\pm$9.6 & 82.3$\pm$8.4 & 87.1$\pm$7.3 \\
     \hline \hline
     \textbf{Our methods} &&&&& \\
     VAE+KNN & 84.4$\pm$4.4 & 76.2$\pm$9.5 & 73.3$\pm$6.7 & 85.9$\pm$4.1 & 90.7$\pm$5.0 \\
     VAE+NN & \textbf{95.5$\pm$4.5} & \textbf{94.4$\pm$5.6} & 69.1$\pm$5.9 & 85.9$\pm$4.1 & 87.0$\pm$8.5 \\
     VAE+AdaBoost & 80.4$\pm$8.5 & 90.0$\pm$10.0 & \textbf{76.0$\pm$4.0} &\textbf{ 89.7$\pm$5.5} & \textbf{90.9$\pm$0.0} \\
	\bottomrule
      \end{tabular}
\end{table*}

\begin{table*} [t] 
  \caption{F-score(\%) and error rate (ER)(\%) of audio event tagging and segmentation tasks; \textit{tag}: audio tagging; \textit{seg}: audio segmentation.}
  \label{audio_table}
\centering
  \begin{tabular}{c|c|c||c|c|c|c}
    \hline
       Task & \multicolumn{2}{c||}{Audio event tagging} & \multicolumn{4}{c}{Audio event segmentation} \\
       Method &
      \multicolumn{2}{c||}{VAE tag} & 
      \multicolumn{2}{c|}{DCASE2017 Baseline} & 
      \multicolumn{2}{c}{VAE tag+VAE seg} \\
    {} & F-score & ER & F-score & ER & F-score & ER \\
    \hline
    Baby cry  & 89.0 & 0.12 & 72.0 & 0.67 & 84.7 & 0.30  \\
    \hline
    Glass break  & 96.0 & 0.04 & 88.5 & 0.22 &  94.1 & 0.12 \\
    \hline
    Gun shot & 85.0 & 0.16 & 57.4 & 0.69 & 87.1 & 0.24  \\
    \hline
    \textbf{Average} & 91.7 & 0.11 & 72.7 & 0.53 & 88.6 & 0.22  \\
    \hline
  \end{tabular}
\end{table*}

\begin{figure*}[t] 
 \centering
 \epsfxsize=110mm \epsffile{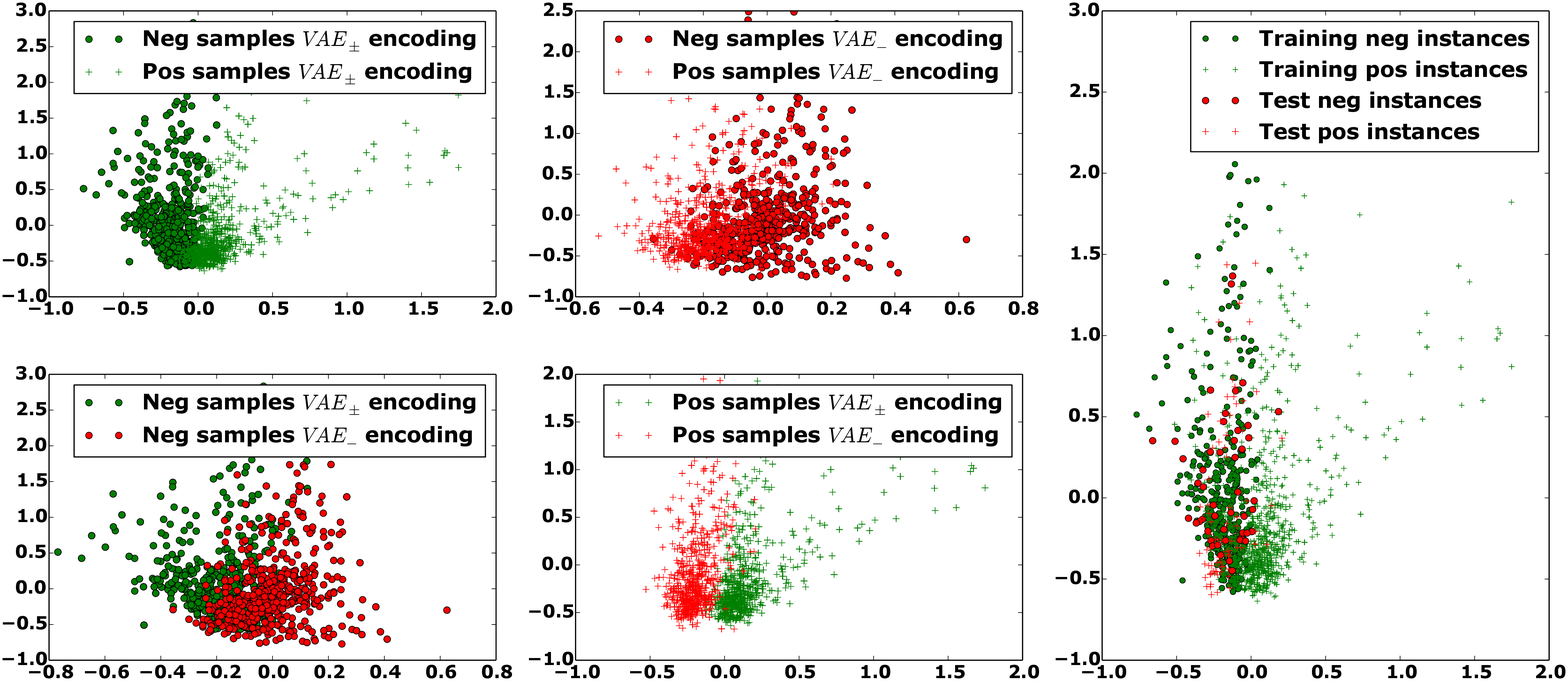}
 \caption{\footnotesize The visualization of the encoding by the trained VAEs  with two latent dimensions. The left four plots show the encoding of positive and negative training instances by both $VAE_{\pm}$ and $VAE_{-}$. Clear separation between the positive and negative instances is observed in the $VAE_{\pm}$ encoding space. The right-most figure shows the encoding of both training and testing data by $VAE_{\pm}$. \label{fig:latent_representation}}
\end{figure*}

\textit{Rare audio event detection dataset}: Audio event detection is another problem that can be formulated as MIL problem where an audio clip and segments within the clip are regarded as the bag and corresponding instances respectively. The goal is to detect whether an audio clip is related to a particular event based on the clip label. We used the rare audio dataset from part of the "Detection and Classification of Acoustic Scenes and Events (DCASE) 2017 Challenge" to evaluate the effectiveness of the proposed method. This dataset contains isolated sound events for 3 target classes: baby cry, gun shot and glass break. Along with recordings of 15 different everyday acoustic scenes as background, including park, coffee shop, bus, street, etc. The audio signal is recorded at 44100 Hz, and downsampled to 22050 Hz as a preprocessing to reduce the computation cost. The target classes and background sounds are synthetically mixed (30 second length) to produce train and test data used in the challenge. The final mixture contains two sets of train and test data each include 500 audio recording per target class (1500 audio files in total). Also the unique event count of each target class in train and test sets are: baby cry-106/42, glass break-96/43, and gun shot-134/53.

We treat each audio recording as a bag and extracted audio features within a moving window as instances. In particular, we used 0.1 and 0.5 second as window size with 50\% overlap. Statistics of thirty-four low level features within the window are regarded as instance features. The thirty-four low level features include Zero crossing rate, energy, spectral centroid, pitch and Mel frequency cepstral coefficients (MFCC) along with their deltas \cite{Davis:1990:CPR:108235.108239}.  These features are extracted from 25 ms frame with 10 ms overlap. The statistic measurements include minimum, maximum, standard deviation, variance, skewness, kurtosis, mean and median. Overall, 272 (34 low level features $\times$ 8 statistics) dimensional features are extracted.  We use random forest based feature selection method to chose the best subset for each class event. Finally, 30 dimensional high level features are used as the system input. In total, we have 500 audio bags for training with 599 or 148 instances (for 0.1 and 0.5 second window, respectively) in each bag.

\subsection{Experiment Results}
\textbf{Comparison} For the MUSK and image annotation benchmark datasets, we follow the previous studies and evaluate the proposed method using 10-fold cross-validation with random fold initialization. Both VAE networks share the same [512, 256, $n_z$, 256, 512] hidden units configuration. The discriminator classifier is a 2-layer network with [64, 64] hidden nodes. Hidden layer size, $n_z$, is chosen experimentally for different datasets, 64 for MUSK1 and Tiger, 32 for Fox, 16 for Elephant and 256 for MUSK2. Through the experiments, we set the batch size to 32 and dropout rate to 0.25. 

As shown in Table \ref{tab:benchmark_results}, the proposed approach achieves significant performance improvement across different benchmark datasets with 4\%, 3.5\%, 12\%, 1.8\% and 3.1\% absolute F-score improvement over MUSK1, MUSK2, Fox, Tiger and Elephant datasets, respectively. Note that some standard deviations in past studies are not available. We have reported the results of our proposed method using 3 different classifiers - k-nearest-neighbor (KNN), Neural Network (NN) and AdaBoost - to investigate the effectiveness of the extracted hidden representations in different classification methods.

For the audio event detection dataset, we compare our results with the DCASE 2017 baseline method. In this dataset, the large data size for each target class is computationally challenging for other MIL approaches, especially the SVM related methods where the training complexity is highly dependent on the size of data. The computational efficiency of the proposed framework comes from the easy training property of the VAE network and the low computational cost in the feature mapping at the test phase. Regardless of the original feature space dimension, the VAE will encode the features into a fixed size vector $n_z$. The experiments highlight the importance of scalability of the MIL method especially in image and audio related applications. Table \ref{audio_table} shows the results for audio event tagging and segmentation tasks, audio event detection with and without time stamps, respectively. Note that DCASE2017 baseline results are only available for the audio event segmentation task at the moment. In fact, with the audio event detection task, the bag level feature lies in 17970 (30*599) dimensional space where the traditional MIL method such as mi-SVM or MI-SVM will take days to train. Using the proposed framework only requires less than half an hour on the same machine with 8*3.7GHz CPUs, one Quadro K5200 GPU and 32GB main memory. 

For audio tagging problem, we used similar network structure as before with the latent dimension $n_z$ set to 128, which gives [512, 256, 128, 256, 512] hidden units configuration. Considering the large dataset, we also changed the batch size to 512. In the testing phase, RBF kernel SVM is chosen experimentally as the final classifier for binary tagging.

For audio segmentation problem, only positive outputs of the audio tagging pipeline are processed with a many-to-many long short term memory (LSTM) network to detect the target events boundaries. At this step, instead of using functional to extract the bag-level representation the 128 dimensional hidden representation of $VAE_{\pm}$ is directly processed with an LSTM network to allow the deep network to self-learn proper features for audio segmentation task. We use two 2-layer LSTM networks with [50, 50] nodes and 0.25 dropout rate. The first LSTM takes the selected 30-dimensional high level audio features summarized in table \ref{audio_table} and the other LSTM takes the $VAE_{\pm}$ 128 dimensional hidden representation as its input. Finally, these two LSTMs are merged together to form a many-to-many output for each audio instance. This network is trained with mean squared error loss function and RMSprop optimization. Experiments have shown the importance of adding the high level audio features along with the VAE representation to improve the segmentation results.

The experimental results confirm the superior performance of the proposed method compared to the baseline system, with average of 91.7\% F-score and 0.11 error rate for audio tagging task and 88.6\% F-score and 0.22 error rate for audio segmentation among three classes. we achieve 15.9\% absolute F-score improvement and 0.31 absolute error rate reduction compared to the DCASE 2017 baseline system. Note that the DCASE 2017 evaluation toolkit is used for audio event segmentation evaluation. We have also compared our proposed segmentation system with a many-to-many-LSTM network trained with high level 30-dimensional audio features described earlier. To have a fair comparison, we don't use the binary tagging results as the input of the segmentation step. Based on the experiments, with a similar network architecture, our proposed solution constantly outperform the LSTM network trained with high level audio feature with F-score absolute improvement of 5.7\%, 6.9\% and 8.5\% for baby cry, glass break and gunshot events.

We stress that the advantage of the proposed framework lies in the scalability of the approach where the high dimension feature can be effectively mapped to low dimension representation through the VAE learning via learning the relations among instances. Approaches like MI-Graph and MI-SVM come with high computational complexity, especially in large dataset with high dimensional features. The low-dimension representation learned with the proposed framework can be used by simple classifier that achieve comparable performance. Moreover, for the audio data, the proposed VAE network automatically learn features at the instance level, thus eliminating the hand-engineered feature extraction such as histogram or Gaussian Mixture Model (GMM) based approaches.

\begin{figure}[t] 
 \centering
 \epsfxsize=60mm \epsffile{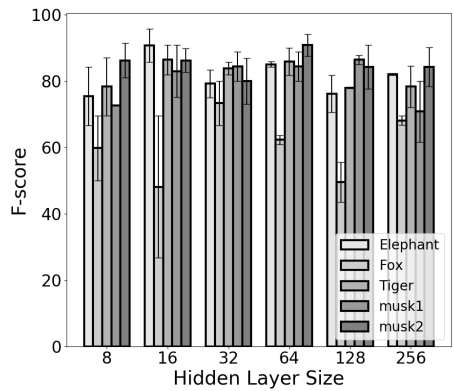}
 \caption{\footnotesize F-score (\%) for MUSK and image annotation datasets using VAE+KNN with different hidden dimensions. \label{fig:hidden_size}}
\end{figure}

\textbf{Visualizations}
In order to have a better insight on what the VAE network learns, we visualize its learned latent representation. Figure \ref{fig:latent_representation} shows the encoding of the training and test data in one of the 10-fold cross-validation evaluation in the Elephant dataset. For visualization purpose, we trained the VAE with two latent dimensions. The encoding of training instances by $VAE_{\pm}$ shows two overlapped data clusters instead of two well separated clusters, corresponding to the positive and negative bag instances. This is as we expected where a positive bag could include negative instances leading to the similar encoding to the negative bag instances. As the counterpart, the $VAE_{-}$ encodes both positive bag instances and negative instances into a single data cluster. As a result, $VAE_{\pm}$ encoding maximizes the difference between the positive and negative instances. More importantly, the similar encoding pattern between positive bag instances and negative bag instances can also be observed in the test data, suggesting effective feature representation of the data.

\textbf{Parameter sensitivity}
We study the sensitivity of the parameter of latent dimension setting $n_z$. Intuitively higher latent dimension setting allows the model to capture more variance of the data, which may over-fit to the data easily when the intrinsic dimension of the problem is low. The lower latent dimension setting emphasizes learning the structure of the data, which may lead to under-fitting in complex problem. As a result, we expect different optimal setting for different problems. This is illustrated in Fig. \ref{fig:hidden_size} where we use the VAE with k-nearest neighbor classifier (VAE+KNN) with different $n_z=[8,16,32,64,128,256]$ on the MUSK and image annotation benchmark datasets. As shown in the figure, the best performance of each tasks is achieved with different $n_z$. For example, for Elephant the best performance is achieved when $n_z=16$ while for MUSK2 $n_z=64$ leads to the best performance. These results follow our intuition, providing a mean to improve the classification performance in different applications.

\section{Conclusion}
In this paper, we presented a novel approach to incorporate deep variational autoencoder into multiple instance learning. We use two VAEs to proximate the posterior of $p(z|X)$ and $p(z|X,Y=-1)$ while applying their latent layers to distinguish the positive bag instances from negative bag instance. The proposed framework also considers the essential challenge of MIL problem where positive label is ambiguous.  Using both theoretical proof and experiments, we have shown that maximizing the distance between two VAEs indeed encourages learning the meaningful representation in the MIL problem. Our experimental results show the scalability and superior performance when compared with the state-of-the-art methods on the benchmark datasets for multiple instance learning task as well as rare audio event detection and segmentation problem. Given the relaxed constrains on data annotation of MIL problem fomulation, the proposed framework can take advantage of the vast amount of weakly labeled data, such as easily available web data, for different applications including image annotation, text categorization, audio event detection, etc. While this paper focuses on binary MIL problem, we are considering various extensions of the proposed framework for other related problems including multi-class MIL and MIML problem. 

\bibliographystyle{plain}
{\footnotesize \bibliography{ref}}
\end{document}